# A New Approach to Intuitionistic Fuzzy Decision Making Based on Projection Technology and Cosine Similarity Measure

Jing Yang[a,*], Wei Su[a]

[a] School of Management and Engineering, Capital University of Economics and Business, Beijing, P.R. China

**Abstract.** For a multi-attribute decision making (MADM) problem, the information of alternatives under different attributes is given in the form of intuitionistic fuzzy number(IFN). Intuitionistic fuzzy set (IFS) plays an important role in dealing with uncertain and incomplete information. The similarity measure of intuitionistic fuzzy sets (IFSs) has always been a research hotspot. A new similarity measure of IFSs based on the projection technology and cosine similarity measure, which considers the direction and length of IFSs at the same time, is first proposed in this paper. The objective of the presented paper is to develop a MADM method and medical diagnosis method under IFS using the projection technology and cosine similarity measure. Some examples are used to illustrate the comparison results of the proposed algorithm and some existing methods. The comparison **result** shows that the proposed algorithm is effective and can identify the optimal scheme accurately. In medical diagnosis area, it can be used to quickly diagnose disease. The proposed method enriches the existing similarity measure methods and it can be applied to not only IFSs, but also other interval-valued intuitionistic fuzzy sets(IVIFSs) as well.

**Keywords**: Fuzzy Multi-attribute Decision Making; Cosine Similarity Measure; Projection Technology; Intuitionistic Fuzzy Set

## 1. Introduction

MADM is one of the most popular research topics in the subject of group decision making at present. Its theory and methods are widely used in engineering, technology, economy, management and military field. In a MADM problem, the optimal alternative is selected or the alternatives are ranked according to multiple attributes. Experts need to provide judgement information for all alternatives when making the decision.

Due to the fuzziness of people's thinking and the complexity of objective things, it is difficult for decision-makers to provide the accurate judgment information. In 1965, an American scholar L.A.Zadeh [1] established the fuzzy set to describe fuzzy phenomena, and proposed the membership function to describe the fuzziness of things. The fuzzy set was a breakthrough of Cantor's classical set theory at the end of the 19th century and laid the foundation of the fuzzy theory.

The membership function in the fuzzy set can only express affirmative information, while the negative information is ignored. Due to the complexity of things, people have difficulties in understanding the uncertainty of things, which makes the traditional fuzzy set challenged in expressing the information.

Atanassov K.T.(1986)[2] put forward the concept of the intuitionistic fuzzy set (IFS) which includes the information of membership and non-membership at the same time. In 1989, Atanassov K.T. [3] extended IFS to the concept of IVIFS. Both IFS and IVIFS can describe the fuzziness of the objective world in details, which are more flexible and practical than traditional fuzzy set in dealing with fuzziness and uncertainty.

[*]Corresponding author. Jing Yang, School of Management and Engineering, Capital University of Economics and Business, 121 Zhang-JiaLuKou, FengTai District, Beijing 100071, P.R.China. E-mail: yjing@cueb.edu.cn.

In the environment of IFS and IVIFS, the proximity degree between the ideal solution and each alternative can be calculated by the cosine similarity measure and the distance measure. And the rank of all alternatives can be determined and the best alternative can be easily identified as well.

Since Atanassov K.T. proposed IFS and IVIFS, the similarity measure and distance measure of IFS and IVIFS have attracted the attention of many scholars and have achieved fruitful results. Distance measure and similarity measure have become the important contents of IFS.

Bustince and Burillo(1995)[4] defined the normalized Hamming distance and normalized Euclidean distance of IFSs and IVIFSs based on membership and non-membership. Szmidt and Kacprzyk (2000)[5] modified the Hamming distance and Euclidean distance by taking the hesitation degree into account.

Chen and Tan (1994)[6] introduced the score function of an IFN. Li Dengfeng and Cheng Chuntian(2002) [7] gave a new similarity measure formula based on the score function. Szmidt and Kacprzyk (2009) [8] proposed a new similarity measure based on the Hausdorff distance between two IFSs.

Xia Liang et al(2013)[9] proposed a new entropy measure with geometrical interpretation of IFSs, which could measure both fuzziness and intuitionism for IFSs. And then they constructed a new similarity measure for IFSs according to the relationship between the entropy and similarity measure.

Beg Ismat and Rashid Tabasam (2016) [10] gave the notion of integral of IFS and introduced an intuitionistic fuzzy implicator and an intuitionistic fuzzy inclusion measure and then proposed a new similarity measure between two IFSs by using an intuitionistic inclusion measure and an intuitionistic fuzzy implication.

Shyi-Ming Chen et al(2016)[11] defined the transformation technique between an IFN and a right-angled triangular fuzzy number, and then proposed a new similarity measure between IFSs based on the centroid points of the transformed right-angled triangular fuzzy number.

Shi Zhan-Hong and Zhang Ding-Hai (2019)[12] believed that a triangular norm could induce an inclusion degree according to the membership and non-membership functions of IFSs. And then they proposed a similarity measure of IFSs by using this triangular norm and the induced inclusion degree.

He Xingxing et al(2019)[13] proposed the concept of intuitionistic fuzzy equivalence, and gave a computational formula for intuitionistic fuzzy equivalencies, which was obtained by combining dissimilarity functions and fuzzy equivalencies. Then they proposed the computational formula for similarity measures on IFSs based on a quaternary function called intuitionistic fuzzy equivalence.

Zhou Lei and Gao Kun (2021)[14] defined the novel pseudometrics on the set of IFNs, which was called the intuitionistic fuzzy cumulative pseudometrics. Then, they presented the unified method to calculate the distances between IFSs, IVIFSs and high-dimensional IFSs based on such pseudometric.

Chen Zichun and Liu Penghui (2022)[15] defined the intuitionistic fuzzy equivalence by using intuitionistic fuzzy negations, intuitionistic fuzzy t-norms, and t-conorms. And then they constructed similarity measures between IFSs by aggregating the intuitionistic fuzzy equivalences.

In addition to the distance measure, the cosine similarity measure has also received attention from researchers. The cosine similarity measure is the cosine of the angle between the vector representations of two IFSs.

Jun Ye(2011)[16] proposed the weighted cosine similarity measure between IFSs and applied it to pattern recognition and medical diagnosis.

Hung and Wang(2012) [17] considered the membership degree, non-membership degree and hesitation degree, and then defined a modified cosine similarity measure for IFSs based on the cosine similarity measure in Jun Ye(2011)[16].

Donghai Liu, Xiaohong Chen, and Dan Peng(2018)[18] studied the cosine similarity measure with hybrid intuitionistic fuzzy information and applied it to medical diagnosis.

Olgun Murat et al (2021) [19] presented a cosine similarity measure for IFSs by using a Choquet integral model in which the interactions between elements are considered.

Some researchers constructed the connection number based on the set pairs analysis(SPA) under environment of IFS, and proposed the Hamming, Euclidean, and Hausdorff distance measures in view of the connection number [20][21][22][23] .

The existing literature mainly focused on the distance measure and the cosine similarity measure of two IFSs. The distance measure considers the distance of two IFSs, and cosine similarity measure only considers the similarity degree in the direction of IFSs, while the similarity degree in the length has been ignored. Therefore, a modified similarity measure based on the cosine similarity measure and projection technology is proposed here, which considers the direction and length of IFSs at the same time. The objective of the presented paper is to develop a MADM method under IFS using the cosine similarity measure and projection technology.

The remainder of this paper is organized as follows: Section 2 introduces the basic concept about IFS, its properties, its score function, and the accuracy function. Section 3 demonstrates some existing similarity measures. Section 4 presents the proposed similarity measure for IFSs. Section 5 describes the application of the proposed method in this paper to decision-making and medical diagnosis. Finally, section 6 concludes the paper.

**2. Basic concepts**

In this part, some basic concepts and definitions about IFS are introduced.

*2.1. IFS*

**Definition 1.** [2] (Atanassov, KT.) Let $U$ be a universe of discourse, then an IFS $A$ in $U$ can be expressed as $A = \{<x, \mu_A(x), \nu_A(x) >| x \in U\}$, where $\mu_A(x)$ represents the membership degree and $\nu_A(x)$ represents the non-membership degree of the element $x$ to the set $U$. For an IFS $A$, $\mu_A(x), \nu_A(x) \in [0,1]$, $0 \le \mu_A(x) + \nu_A(x) \le 1$ for $\forall x \in U$.

For an IFS $A$ in $U$, let $\pi_A(x) = 1 - \mu_A(x) - \nu_A(x)$, then $\pi_A(x)$ is called the hesitancy degree of the element $x$ to $A$.

Obviously, $0 \le \pi_A(x)) \le 1$.

**Property 1.**
If there are two IFSs $A$ and $B$, their relations can be defined as follows[2] (Atanassov and Gargov 1986):
① $A = B$ if and only if $\mu_A(x) = \mu_B(x)$, $\nu_A(x) = \nu_B(x)$ for any $x$ in U;
② $A \subseteq B$ if and only if $\mu_A(x) \le \mu_B(x)$, $\nu_A(x) \ge \nu_B(x)$ for any $x$ in U;

**Property 2.** [24]
Let $a = (\mu_a, \nu_a)$ $b = (\mu_b, \nu_b)$ be two intuitionistic fuzzy numbers (IFNs), then
① $a \cap b = (\min\{\mu_a, \mu_b\}, \max\{\nu_a, \nu_b\})$
② $a \cup b = (\max\{\mu_a, \mu_b\}, \min\{\nu_a, \nu_b\})$
③ $a \oplus b = (\mu_a + \mu_b - \mu_a \cdot \mu_b, \nu_a \cdot \nu_b)$
④ $a \otimes b = (\mu_a \cdot \mu_b, \nu_a + \nu_b - \nu_a \cdot \nu_b)$
⑤ $\lambda a = (1 - (1 - \mu_a)^\lambda, (\nu_a)^\lambda)$, $\lambda > 0$
⑥ $a^\lambda = ((\mu_a)^\lambda, 1 - (1 - \nu_a)^\lambda)$, $\lambda > 0$

**Definition 2.** [25] (Xu, Z.S 2006)
Let $a = (\mu_a, \nu_a)$ be an IFN, then the score function $s(a)$ and the accuracy function $h(a)$ can be defined as follows:

$$s(a) = \mu_a - \nu_a \qquad (1)$$
$$h(a) = \mu_a + \nu_a \qquad (2)$$

Obviously, the larger $s(a)$ is, the larger $a$ is.

**Property 3.** [25] (Xu, Z.S 2006)
If there are two IFNs $a$ and $b$, $s(a)$ and $s(b)$ are their score functions, $h(a)$ and $h(b)$ are their accuracy functions. If $a > b$ exists, one of the following two conditions must be met:
① $s(a) > s(b)$;
② $s(a) = s(b)$ and $h(a) > h(b)$.

For example, if $a = (0.3, 0.1)$ and $b = (0.4, 0.2)$, then $s(a) = 0.2$, $s(b) = 0.2$. So, the two IFNs $a$ and $b$ cannot be compared according to the score function. The accuracy function can make up for this deficiency.

According to equation(2), $h(a) = 0.4$, $h(b) = 0.6$, $h(a) < h(b)$. Therefore, $a < b$.

## 3. Some existing similarity measures

In this section, some existing similarity measures of intuitionistic fuzzy sets are introduced, which include the distance measure, cosine similarity measure and similarity measure based on the set pair analysis.

### 3.1 The similarity measures based on the distance

In a universe of discourse $U = \{x_1, x_2, \cdots, x_n\}$, suppose there are two IFSs $A$ and $B$:

$$A = \{<x_i, \mu_A(x_i), \nu_A(x_i)> | x_i \in U\}$$
$$B = \{<x_i, \mu_B(x_i), \nu_B(x_i)> | x_i \in U\}$$
$$i = 1, 2, \cdots, n$$

Bustince and Burillo (1995)[4] proposed the following normalized Hamming distance.

$$d_1(A, B) = \frac{1}{2}\sum_{i=1}^{n} w_i(|\mu_A(x_i) - \mu_B(x_i)| + |\nu_A(x_i) - \nu_B(x_i)|)$$

$$d_2(A, B) = \frac{1}{2}\sum_{i=1}^{n} w_i[(\mu_A(x_i) - \mu_B(x_i))^2 + (\nu_A(x_i) - \nu_B(x_i))^2]$$

Based on these distance measures, the similarity measures can be calculated as follows:

$$S_1(A, B) = 1 - d_1(A, B) = 1 - \frac{1}{2}\sum_{i=1}^{n} w_i(|\mu_A(x_i) - \mu_B(x_i)| + |\nu_A(x_i) - \nu_B(x_i)|) \tag{3}$$

$$S_2(A, B) = 1 - d_2(A, B) = 1 - \frac{1}{2}\sum_{i=1}^{n} w_i[(\mu_A(x_i) - \mu_B(x_i))^2 + (\nu_A(x_i) - \nu_B(x_i))^2] \tag{4}$$

where $w_i$ is the weight of $x_i$, $\sum_{i=1}^{n} w_i = 1$.

Szmidt and Kacprzyk (2000)[5])gave the following two distance measure formulas based on the distance measure in Bustince and Burillo(1995)[4], taking into account the membership degree, non-membership degree and hesitation degree.

$$d_3(A, B) = \frac{1}{2}\sum_{i=1}^{n} w_i(|\mu_A(x_i) - \mu_B(x_i)| + |\nu_A(x_i) - \nu_B(x_i)| + |\pi_A(x_i) - \pi_B(x_i)|)$$

$$d_4(A, B) = \frac{1}{2}\sum_{i=1}^{n} w_i[(\mu_A(x_i) - \mu_B(x_i))^2 + (\nu_A(x_i) - \nu_B(x_i))^2 + (\pi_A(x_i) - \pi_B(x_i))^2]$$

where $\pi_A(x_i) = 1 - \mu_A(x_i) - \nu_A(x_i)$, $\pi_B(x_i) = 1 - \mu_B(x_i) - \nu_B(x_i)$,

Based on these distance measures, the similarity measures can be calculated as follows:

$$S_3(A, B) = 1 - d_3(A, B) = 1 - \frac{1}{2}\sum_{i=1}^{n} w_i(|\mu_A(x_i) - \mu_B(x_i)| + |\nu_A(x_i) - \nu_B(x_i)| + |\pi_A(x_i) - \pi_B(x_i)|) \tag{5}$$

$$S_4(A, B) = 1 - d_4(A, B) = 1 - \frac{1}{2}\sum_{i=1}^{n} w_i[(\mu_A(x_i) - \mu_B(x_i))^2 + (\nu_A(x_i) - \nu_B(x_i))^2 + (\pi_A(x_i) - \pi_B(x_i))^2] \tag{6}$$

Szmidt and Kacprzyk(2009)[8] proposed the following Hausdorff distance between two IFSs.

$$d_5(A, B) = \sum_{i=1}^{n} w_i \max\{|\mu_A(x_i) - \mu_B(x_i)|, |\nu_A(x_i) - \nu_B(x_i)|, |\pi_A(x_i) - \pi_B(x_i)|\}$$

Then, the similarity measure can be calculated as follows:

$$S_5(A, B) = 1 - d_5(A, B) = 1 - \sum_{i=1}^{n} w_i \max\{|\mu_A(x_i) - \mu_B(x_i)|, |\nu_A(x_i) - \nu_B(x_i)|, |\pi_A(x_i) - \pi_B(x_i)|\} \tag{7}$$

Li Dengfeng and Cheng Chuntian(2002)[7] gave the following distance measure formula

$$d_6(A,B) = \sqrt[p]{\sum_{i=1}^{n} w_i |\varphi_A(x_i) - \varphi_B(x_i)|^p}$$

where $1 \leq p \leq \infty$, $\varphi_A(x_i) = \dfrac{\mu_A(x_i) + 1 - \nu_A(x_i)}{2}$, $\varphi_B(x_i) = \dfrac{\mu_B(x_i) + 1 - \nu_B(x_i)}{2}$.

Based on this distance measure, the similarity measure can be calculated as follows:

$$S_6(A,B) = 1 - d_6(A,B) = 1 - \sqrt[p]{\sum_{i=1}^{n} w_i |\varphi_A(x_i) - \varphi_B(x_i)|^p} \tag{8}$$

The above distance formulas express the distance between two IFSs $A$ and $B$. The smaller the distance is, the more similar $A$ and $B$ are.

### 3.2 The similarity measures based on the cosine similarity

Here are some cosine similarity measures.

The cosine similarity measures for IFSs in Jun Ye (2011)[16] is defined as follows,

$$S_7(A,B) = \sum_{i=1}^{n} w_i \frac{\mu_A(x_i) \cdot \mu_B(x_i) + \nu_A(x_i) \cdot \nu_B(x_i)}{\sqrt{(\mu_A(x_i))^2 + (\nu_A(x_i))^2} \cdot \sqrt{(\mu_B(x_i))^2 + (\nu_B(x_i))^2}} \tag{9}$$

Hung and Wang (2012)[17] pointed out some drawbacks of the cosine similarity measure in Jun Ye (2011)[16] and defined a modified cosine similarity measure for IFSs as follows:

$$S_8(A,B) = \sum_{i=1}^{n} w_i \frac{\mu_A(x_i) \cdot \mu_B(x_i) + \nu_A(x_i) \cdot \nu_B(x_i) + \pi_A(x_i) \cdot \pi_B(x_i)}{\sqrt{(\mu_A(x_i))^2 + (\nu_A(x_i))^2 + (\pi_A(x_i))^2} \cdot \sqrt{(\mu_B(x_i))^2 + (\nu_B(x_i))^2 + (\pi_B(x_i))^2}} \tag{10}$$

### 3.3 The similarity measures based on the set pair analysis

Here are some similarity measures based on the set pair analysis.

The similarity measure for IFSs in Harish Garg and Kamal Kumar (2018)[13] is defined as follows,

$$S_9(A,B) = \sum_{i=1}^{n} w_i (1 - \frac{|a_A(x_i) - a_B(x_i)| + |b_A(x_i) - b_B(x_i)| + |c_A(x_i) - c_B(x_i)|}{3}) \tag{11}$$

Another similarity measure for IFSs in Harish Garg and Kamal Kumar (2018)[13] is defined as follows,

$$S_{10}(A,B) = \frac{\sum_{i=1}^{n} w_i \{\min[a_A(x_i), a_B(x_i)] + \min[b_A(x_i), b_B(x_i)] + \min[c_A(x_i), c_B(x_i)]\}}{\sum_{i=1}^{n} w_i \{\max[a_A(x_i), a_B(x_i)] + \max[b_A(x_i), b_B(x_i)] + \max[c_A(x_i), c_B(x_i)]\}} \tag{12}$$

Where, $a(x_i) = \mu(x_i) \cdot (1 - \nu(x_i))$, $c(x_i) = \nu(x_i) \cdot (1 - \mu(x_i))$ and $b(x_i) = 1 - \mu(x_i) \cdot (1 - \nu(x_i)) - \nu(x_i) \cdot (1 - \mu(x_i))$ in equation (11) and (12).

## 4. The proposed similarity measure for IFSs

In this section, the membership degree, non-membership degree and hesitancy degree of IFS based on the cosine similarity measure in Hung and Wang (2012)[17]) are first introduced, and then a novel similarity measure based on cosine similarity measure and projection technology is proposed here, which considers cosine similarity measure and projection technology at the same time.

### 4.1 Cosine Similarity measure for IFSs

The cosine similarity measure is defined as the inner product of two vectors divided by the product of their lengths. It is the cosine of the angle between the vector representations of two fuzzy sets.

If $a$ and $b$ are two IFNs, the cosine similarity measure $C(a,b)$ between $a$ and $b$ can be expressed as follows:

$$a = (\mu_a, \nu_a), b = (\mu_b, \nu_b)$$

$$C(a,b) = \frac{\mu_a \cdot \mu_b + \nu_a \cdot \nu_b + \pi_a \cdot \pi_b}{\sqrt{(\mu_a^2 + \nu_a^2 + \pi_a^2)} \cdot \sqrt{\mu_b^2 + \nu_b^2 + \pi_b^2}} \tag{13}$$

where $\pi_a = 1 - \mu_a - \nu_a$, $\pi_b = 1 - \mu_b - \nu_b$

Obviously, the cosine similarity measurement $C(a,b)$ between $a$ and $b$ satisfies the following properties:

**Property 4.**

① $0 \leq C(a,b) \leq 1$

② $C(a,b) = 1$ if and only if $a = b$

③ $C(a,b) = C(b,a)$

**Proof.**

The inequality in property4① $C(a,b) \geq 0$ is obvious. The inequality $C(a,b) \leq 1$ in property 4① can be proved as follows:

$$\begin{pmatrix} (\mu_a \nu_b - \nu_a \mu_b)^2 \geq 0 \\ (\mu_a \pi_b - \pi_a \mu_b)^2 \geq 0 \\ (\nu_a \pi_b - \pi_a \nu_b)^2 \geq 0 \end{pmatrix}$$

$$\Rightarrow (\mu_a \nu_b - \nu_a \mu_b)^2 + (\mu_a \pi_b - \pi_a \mu_b)^2 + (\nu_a \pi_b - \pi_a \nu_b)^2 \geq 0$$

$$\Rightarrow \mu_a^2 \nu_b^2 + \nu_a^2 \mu_b^2 + \mu_a^2 \pi_b^2 + \pi_a^2 \mu_b^2 + \nu_a^2 \pi_b^2 + \pi_a^2 \nu_b^2 \geq 2\mu_a \mu_b \nu_a \nu_b + 2\mu_a \pi_a \mu_b \pi_b + 2\nu_a \pi_a \nu_b \pi_b$$

$$\Rightarrow \mu_a^2 \mu_b^2 + \pi_a^2 \pi_b^2 + \nu_a^2 \nu_b^2 + \mu_a^2 \nu_b^2 + \nu_a^2 \mu_b^2 + \mu_a^2 \pi_b^2 + \pi_a^2 \mu_b^2 + \nu_a^2 \pi_b^2 + \pi_a^2 \nu_b^2$$
$$\geq \mu_a^2 \mu_b^2 + \pi_a^2 \pi_b^2 + \nu_a^2 \nu_b^2 + 2\mu_a \mu_b \nu_a \nu_b + 2\mu_a \pi_a \mu_b \pi_b + 2\nu_a \pi_a \nu_b \pi_b$$

$$\Rightarrow (\mu_a^2 + \nu_a^2 + \pi_a^2) \cdot (\mu_b^2 + \nu_b^2 + \pi_b^2) \geq (\mu_a \mu_b + \nu_a \nu_b + \pi_a \pi_b)^2$$

$$\Rightarrow C(a,b) = \frac{\mu_a \cdot \mu_b + \nu_a \cdot \nu_b + \pi_a \cdot \pi_b}{\sqrt{(\mu_a^2 + \nu_a^2 + \pi_a^2)} \cdot \sqrt{\mu_b^2 + \nu_b^2 + \pi_b^2}} \leq 1$$

Property4② and property4③ are straightforward.

□

Next, the cosine similarity measure between IFNs can be extended to IFSs. In a universe of discourse $U = \{x_1, x_2, \cdots, x_n\}$, let $A$ and $B$ be two IFSs, then $A$ and $B$ can be expressed as follows:

$$A = \{< x_i, \mu_A(x_i), \nu_A(x_i) > | x_i \in U\}$$
$$B = \{< x_i, \mu_B(x_i), \nu_B(x_i) > | x_i \in U\} \quad i = 1, 2, \cdots, n$$

The cosine similarity measure between $A$ and $B$ can be defined as follows:

$$C(A,B) = \sum_{i=1}^{n} w_i \frac{\mu_A(x_i) \cdot \mu_B(x_i) + \nu_A(x_i) \cdot \nu_B(x_i) + \pi_A(x_i) \cdot \pi_B(x_i)}{\sqrt{(\mu_A(x_i))^2 + (\nu_A(x_i))^2 + (\pi_A(x_i))^2} \cdot \sqrt{(\mu_B(x_i))^2 + (\nu_B(x_i))^2 + (\pi_B(x_i))^2}} \tag{14}$$

Where $w_i$ means the weight of each element $x_i (i=1,2,\cdots,n)$, $0 \leq w_i \leq 1$, $\sum_{i=1}^{n} w_i = 1$.

In the same way, the cosine similarity measure $C(A,B)$ between $A$ and $B$ has the following properties as well:
  ①  $0 \leq C(A,B) \leq 1$
  ②  $C(A,B) = 1$ if and only if $A = B$
  ③  $C(A,B) = C(B,A)$

*4.2 A novel similarity measure for IFSs based on cosine similarity measure and projection technology*

Cosine similarity measure only considers the similarity degree of IFSs in the direction, while the similarity degree in the length has been ignored. Therefore, a novel similarity measure based on cosine similarity measure and projection technology is proposed here, which considers cosine similarity measure and projection technology at the same time. The proposed similarity measure of IFSs $A$ on $B$ is given below.

$$\begin{aligned}S(A,B) &= \sum_{i=1}^{n} w_i \sqrt{(\mu_A(x_i))^2 + (v_A(x_i))^2 + (\pi_A(x_i))^2} \cdot C(A,B) \\ &= \sum_{i=1}^{n} w_i \frac{\mu_A(x_i) \cdot \mu_B(x_i) + v_A(x_i) \cdot v_B(x_i) + \pi_A(x_i) \cdot \pi_B(x_i)}{\sqrt{(\mu_B(x_i))^2 + (v_B(x_i))^2 + (\pi_B(x_i))^2}}\end{aligned} \qquad (15)$$

Obviously, the larger $S(A,B)$ is, the more similar $A$ and $B$ is.

## 5. The application of the proposed similarity measure in MADM and medical diagnosis

The proposed similarity measure in this paper can be used in many fields, such as decision-making, pattern recognition and medical diagnosis etc.

*5.1 The application of the proposed similarity measure in MADM*

In the process of decision-making, the ideal IFS will be set first, and then each scheme will be compared with the optimal scheme. The greater the degree of similarity is, the better the corresponding scheme is.

5.1.1 An ideal IFS

In the decision-making process, the decision-maker will consider a number of influencing factors for different alternatives, which are referred to as indicators. Indicators are divided into benefit indicators and cost indicators. For benefit indicators, such as profit and rate of capital return, the greater the value is, the better the scheme is. For cost indicators, such as investment risk, investment amount, and maintenance cost, the smaller the value is, the better the scheme is. The decision-maker has to identify the best alternative according to these indicators. Suppose the set of alternatives is represented by $A$, the set of indicators is represented by $C$. There are two categories of indicators: benefit indicators and cost indicators. The set of benefit indicators is represented by $C_B$, and the set of cost indicators is represented by $C_C$. The value of an indicator in the optimal alternative can be expressed by IFN.

For benefit indicators, the value of a benefit indicator can be calculated as follows:

$$r_j^* = (\mu_j^*, v_j^*) = (\max_i(\mu_{ij}), \min_i(v_{ij})) \qquad (16)$$

Where $i$ is the index of the set of alternatives $A$, and $j$ is the index of the set of benefit indicators $C_B$;

For cost indicators, the value of a cost indicator can be calculated as follows:

$$r_j^* = (\mu_j^*, v_j^*) = (\min_i(\mu_{ij}), \max_i(v_{ij})) \qquad (17)$$

Where $i$ is the index of the set of alternatives $A$, and $j$ is the index of the set of benefit indicators $C_C$;

The score function of the ideal IFN can be described by $s(r_j^*)$.

$s(r_j^*) = \mu_j^* - v_j^*$, where $j$ is the index of the set of indicators $C$;

5.1.2 Intuitionistic fuzzy decision making based on the proposed method

In this part, how to find the best solution among multiple alternatives under intuitionistic fuzzy environment is studied.

Let $A = \{A_1, A_2, \cdots, A_n\}$ be a set of alternatives, and $C = \{C_1, C_2, \cdots, C_m\}$ be a set of indicators. There are two categories of indicators: benefit indicators and cost indicators. The set of benefit indicators is represented by $C_B$, and the set of cost indicators is represented by $C_C$. The original data matrix can be expressed as follows:

$$R = \begin{pmatrix} r_{11} & r_{12} & \cdots & r_{1m} \\ r_{21} & r_{22} & \cdots & r_{2m} \\ \vdots & \vdots & \vdots & \vdots \\ r_{n1} & r_{n2} & \cdots & r_{nm} \end{pmatrix}$$

where $r_{ij} = (\mu_{ij}, v_{ij})$, $0 \leq \mu_{ij}, v_{ij} \leq 1$, $0 \leq \mu_{ij} + v_{ij} \leq 1$, $0 \leq i \leq n$ and $0 \leq j \leq m$.

There are $m$ indexes in each alternative to represent its characteristics, which is:
$A_i = (r_{i1}, r_{i2}, \cdots, r_{im}) = ((\mu_{i1}, v_{i1}), (\mu_{i2}, v_{i2}), \cdots, (\mu_{im}, v_{im}))$, where $0 \leq i \leq n$.

The following are the steps of the decision-making process.

Step1: set an ideal optimal scheme.

In the process of decision making, an ideal optimal scheme can be decided according to equation (16) and (17). The ideal optimal scheme is a combination of the optimal value in each index.

For benefit indicators, the ideal IFN can be expressed as follows:
$r_j^* = (\mu_j^*, v_j^*) = (\max_i(\mu_{ij}), \min_i(v_{ij}))$, where $j$ is the index of the set of benefit indicators $C_B$.

For cost indicators, the ideal IFN can be expressed as follows:
$r_j^* = (\mu_j^*, v_j^*) = (\min_i(\mu_{ij}), \max_i(v_{ij}))$, where $j$ is the index of the set of benefit indicators $C_C$.

The ideal optimal scheme can be expressed by $A^*$.
$A^* = (r_1^*, r_2^*, \cdots, r_m^*) = ((\mu_1^*, v_1^*), (\mu_2^*, v_2^*), \cdots, (\mu_m^*, v_m^*))$

Step2: calculate the similarity measure.

Then, each scheme can be compared with the ideal optimal one, and their similarity can be calculated. The similarity measure $S(A_i, A^*)$ between $A_i$ and $A^*$ can be calculated as follows:

$$S(A_i, A^*) = \sum_{j=1}^{m} w_j \sqrt{(\mu_{ij})^2 + (v_{ij})^2 + (\pi_{ij})^2} \cdot C(A_i, A^*)$$
$$= \sum_{j=1}^{m} w_j \frac{\mu_{ij} \cdot \mu_j^* + v_{ij} \cdot v_j^* + \pi_{ij} \cdot \pi_j^*}{\sqrt{(\mu_j^*)^2 + (v_j^*)^2 + (\pi_j^*)^2}}$$

(18)

where $\pi_{ij} = 1 - \mu_{ij} - v_{ij}$, $\pi_j^* = 1 - \mu_j^* - v_j^*$, $w_j$ means the weight of indicators, $i = 1, 2, \cdots, n$, and $j = 1, 2, \cdots, m$.

Step3: choose the best alternative.

Finally, the alternatives will be ranked according to the similarity degree. The greater the degree of similarity is, the better the corresponding scheme is.

*5.1.3 A Numerical Example and Comparative Analysis*

*5.1.3.1 Numerical Example*

In this part, an example 【13】 is used to verify the effectiveness of the proposed method.

Supply chain management refers to the entire process of optimizing the operation of the supply chain, from the beginning of procurement to the satisfaction of end customers, with minimal cost. Supply chain management emphasizes that upstream enterprises maintain a good cooperative relationship with downstream enterprises, which can reduce total costs and inventory, achieve rapid response, and enhance competitive advantage. Therefore, for enterprises, they will carefully choose their suppliers.

Suppose there is a company who has five suppliers to choose from: $A_1$, $A_2$, $A_3$, $A_4$ and $A_5$. The following indicators need to be considered when selecting a supplier: (1) $C_1$ is the supply capacity; (2) $C_2$ is the product quality; (3) $C_3$ is the product price; (4) $C_4$ is the service level. Where $C_1$, $C_2$ and $C_4$ are benefit indicators and $C_3$ is the cost indicator. These four indicators are not equally important in decision making, so they are given different weights. The weight vector of the four indicators is as follows: $w = (0.3, 0.3, 0.2, 0.2)$.

Step 1: Establish the decision matrix.

After investigating and evaluating various suppliers, these five suppliers are evaluated under the above four indicators by the form of IFSs, as shown in Table 1.

Table 1 Decision Matrix

| Suppliers | $C_1$ | $C_2$ | $C_3$ | $C_4$ |
|---|---|---|---|---|
| $A_1$ | (0.8,0.1) | (0.5,0.2) | (0.5,0.3) | (0.5,0.2) |
| $A_2$ | (0.5,0.2) | (0.6,0.2) | (0.6,0.2) | (0.6,0.1) |
| $A_3$ | (0.3,0.2) | (0.8,0.1) | (0.6,0.2) | (0.8,0.1) |
| $A_4$ | (0.7,0.1) | (0.6,0.2) | (0.5,0.4) | (0.7,0.1) |
| $A_5$ | (0.6,0.2) | (0.5,0.4) | (0.8,0.1) | (0.4,0.3) |

Step 2: Decide the ideal optimal supplier.

$A^* = ((0.8,0.1), (0.8,0.1), (0.5,0.4), (0.8,0.1))$

Step 3: Calculate the similarity degree.

According to equation (18), the similarity measure $S(A_i, A^*)$ between $A_i$ and $A^*$ can be calculated.

$S(A_1, A^*) = 0.6411$

$S(A_2, A^*) = 0.6096$

$S(A_3, A^*) = 0.6441$

$S(A_4, A^*) = 0.6847$

$S(A_5, A^*) = 0.5906$

Step 4: Choose the best supplier.

According to the degree of similarity, the suppliers can be ranked as follows. The more similar the supplier is to the ideal optimal scheme, the better the supplier is.

$S(A_4, A^*) > S(A_3, A^*) > S(A_1, A^*) > S(A_2, A^*) > S(A_5, A^*)$

The priority of suppliers can be decided as follows:

$A_4 \succ A_3 \succ A_1 \succ A_2 \succ A_5$

Therefore, the best alternative should be $A_4$.

*5.1.3.2 Comparative Analysis*

To verify the feasibility of the proposed method, the result of the proposed algorithm is compared with results of other algorithms in literature [4],[7],[13] and [16] with the same weight $w = (0.3, 0.3, 0.2, 0.2)$.

The comparison result is shown in Table 2.

Table 2 Comparison Result of Different Algorithms

| Author | Method | The similarity measure | Best Supplier |
|---|---|---|---|

| Bustince and Burillo (1995) | The similarity measure based on normalized Hamming distance [4] | $S_2(A_1, A^*) = 0.974$ $S_2(A_2, A^*) = 0.9725$ $S_2(A_3, A^*) = 0.956$ $S_2(A_4, A^*) = 0.99$ $S_2(A_5, A^*) = 0.9275$ | $A_4$ |
|---|---|---|---|
| Li Dengfeng and Cheng Chuntian(2002) | Li Dengfeng's distance measure [7]) ( $p=2$ ) | $S_6(A_1, A^*) = 0.8568$ $S_6(A_2, A^*) = 0.8411$ $S_6(A_3, A^*) = 0.8225$ $S_6(A_4, A^*) = 0.9106$ $S_6(A_5, A^*) = 0.7359$ | $A_4$ |
| Jun Ye(2011) | Jun Ye's cosine similarity measure [16] | $S_7(A_1, A^*) = 0.9819$ $S_7(A_2, A^*) = 0.9719$ $S_7(A_3, A^*) = 0.9561$ $S_7(A_4, A^*) = 0.9940$ $S_7(A_5, A^*) = 0.9809$ | $A_4$ |
| Harish Garg and Kamal Kumar (2018) | Harish Garg and Kamal Kumar's similarity measure based on the set pair analysis [13] | $S_9(A_1, A^*) = 0.8867$ $S_9(A_2, A^*) = 0.8473$ $S_9(A_3, A^*) = 0.88$ $S_9(A_4, A^*) = 0.922$ $S_9(A_5, A^*) = 0.7533$ | $A_4$ |
| Author | The proposed method in this paper | $S(A_1, A^*) = 0.6411$ $S(A_2, A^*) = 0.6096$ $S(A_3, A^*) = 0.6441$ $S(A_4, A^*) = 0.6847$ $S(A_5, A^*) = 0.5906$ | $A_4$ |

As what is shown in Table 2, the best supplier of the proposed algorithm is consistent with the result of other existing methods, and the best one is $A_4$. Bustince and Burillo(1995) proposed normalized Hamming distance. Li Deng feng and Cheng Chun tian(2002) proposed the similarity measure based on the distance measure. Jun Ye(2011) extended the concept of the cosine similarity measure between fuzzy sets to a weighted cosine similarity measure between IFSs. Harish Garg and Kamal Kumar (2018) proposed a novel similarity measure to measure the relative strength of the different IFSs by using the connection number, which was the main component of the set pair analysis theory.

These authors have studied the similarity of IFSs from different perspectives such as the distance, the angle cosine and the set pair analysis theory. The analysis methods are different, but the final conclusions are the same. This paper proposes a similarity measure method based on the cosine similarity and projection technology, in which both the similarity in direction and the similarity in length have been considered in calculating the similarity of IFSs. In this way, more information can be included.

*5.2 The application of the proposed similarity measure in medical diagnosis*

Medical diagnosis refers to the process of finding out the location and extent of the disease and determining the name of the disease when the human body is in an abnormal state. When a patient sees a doctor, he will first describe his symptoms, and the doctor will judge what kind of disease the patient suffers from by comparing the symptoms with different diseases. For example, when a person vomits, the doctor will judge whether he has gastroenteritis, peptic ulcer, pyloric obstruction, acute gastric dilatation or functional dyspepsia according to the patient's description and laboratory indicators.

Medical diagnosis has two parts: one is a number of known standard models with some characteristics, and the other is the object to be identified. In short, medical diagnosis is to recognize and classify the research object according to some characteristics.

5.2.1 Algorithms for medical diagnosis

The method of medical diagnosis is as follows:

Let $A = \{A_1, A_2, \cdots, A_n\}$ be a set of possible diseases, and $C = \{C_1, C_2, \cdots, C_m\}$ be a set of indicators, including typical symptoms, signs and examination results. $B = \{b_1, b_2, \cdots, b_m\}$ represents the patient. The original data matrix can be expressed as follows:

$$R = \begin{pmatrix} r_{11} & r_{12} & \cdots & r_{1m} \\ r_{21} & r_{22} & \cdots & r_{2m} \\ \vdots & \vdots & \vdots & \vdots \\ r_{n1} & r_{n2} & \cdots & r_{nm} \\ b_1 & b_2 & \cdots & b_m \end{pmatrix}$$

where $r_{ij} = (\mu_{ij}, v_{ij})$, $b_j = (\mu_{bj}, v_{bj})$, $0 \leq i \leq n$ and $0 \leq j \leq m$.

There are $m$ indexes in each alternative to represent its characteristics, which is:

$A_i = (r_{i1}, r_{i2}, \cdots, r_{im}) = ((\mu_{i1}, v_{i1}), (\mu_{i2}, v_{i2}), \cdots, (\mu_{im}, v_{im}))$

Then, each scheme $A_i$ can be compared with $B$ and the similarity measure $S(A_i, B)$ between $A_i$ and $B$ can be calculated as follows:

$$S(A_i, B) = \sum_{j=1}^{m} \sqrt{(\mu_{bj})^2 + (v_{bj})^2 + (\pi_{bj})^2} \cdot C(A_i, B) = \sum_{j=1}^{m} \frac{\mu_{ij} \cdot \mu_{bj} + v_{ij} \cdot v_{bj} + \pi_{ij} \cdot \pi_{bj}}{\sqrt{(\mu_{ij})^2 + (v_{ij})^2 + (\pi_{ij})^2}} \qquad (19)$$

where $\pi_{ij} = 1 - \mu_{ij} - v_{ij}$, $\pi_{bj} = 1 - \mu_{bj} - v_{bj}$, $i = 1, 2, \cdots, n$ and $j = 1, 2, \cdots, m$

Let $S(A_{i_0}, B) = \max S(A_i, B)$, then the patient has $A_{i_0}$ disease.

5.2.2 *Numerical Example*

In this part, a numerical example adapted from the literature ([26] I.K. Vlachos, G.D. Sergiadis) is used to prove the effectiveness of the above method.

Suppose there is a set of diseases $A = \{A_1(\textit{Viral fever}), A_2(\textit{Malaria}), A_3(\textit{Typhoid}), A_4(\textit{Stomach problem}), A_5(\textit{Chest problem})\}$ and $C = \{C_1(\textit{Temperature}), C_2(\textit{Headache}), C_3(\textit{Stomach pain}), C_4(\textit{Cough}), C_5(\textit{Chest pain})\}$ represent a set of symptoms(indicators). Each disease has several symptoms. There is a patient whose symptoms are described as follows: $B = \{<C_1, 0.8, 0.1>, <C_2, 0.6, 0.1>, <C_3, 0.2, 0.8>, <C_4, 0.6, 0.1>, <C_5, 0.1, 0.6>\}$. The value of a symptom in disease can be expressed by IFN. The original data matrix can be expressed as follows:

Table 3  The original data matrix

| Diseases | $C_1$ | $C_2$ | $C_3$ | $C_4$ | $C_5$ |
|---|---|---|---|---|---|
| $A_1$ | <0.4,0.0> | <0.3,0.5> | <0.1,0.7> | <0.4,0.3> | <0.1,0.7> |
| $A_2$ | <0.7,0.0> | <0.2,0.6> | <0.0,0.9> | <0.7,0.0> | <0.1,0.8> |
| $A_3$ | <0.3,0.3> | <0.6,0.1> | <0.2,0.7> | <0.2,0.6> | <0.1,0.9> |
| $A_4$ | $<0.1,0.7>$ | $<0.2,0.4>$ | $<0.8,0.0>$ | $<0.2,0.7>$ | $<0.2,0.7>$ |
| $B$ | $<0.8,0.1>$ | $<0.6,0.1>$ | $<0.2,0.8>$ | $<0.6,0.1>$ | $<0.1,0.6>$ |

The aim is to judge what kind of diseases the patient in suffering from according to his symptoms.

Each disease $A_i$ can be compared with $B$, and the similarity measure $S(A_i, B)$ between $A_i$ and $B$ can be calculated as follows:

$S(A_1, B) = 2.6628$

$S(A_2, B) = 2.6674$

$S(A_3, B) = 2.4442$

$S(A_4, B) = 1.4949$

According to the similarity measure $S(A_i, B)$, the patient is suffering from the disease $A_2$ (*Malaria*).

*5.2.3 The result Comparative Analysis*

To verify the feasibility of the proposed method, the result of the proposed algorithm is compared with results of other algorithms in literature [4],[5],[8],[16]and [17] with the weight $w = (0.2, 0.2, 0.2, 0.2, 0.2)$.

The comparison result is shown in Table 4.

Table4  Comparison Result of Different Algorithms

| Author | Method | The similarity measure | The patient's disease |
|---|---|---|---|
| Bustince and Burillo (1995) | The similarity measure based on normalized Hamming distance [4] | $S_1(A_1, B) = 0.81$<br>$S_1(A_2, B) = 0.82$<br>$S_1(A_3, B) = 0.80$<br>$S_1(A_4, B) = 0.54$ | $A_2$ |
| Szmidt and Kacprzyk (2000) | Szmidt and Kacprzyk's the similarity measure [5] | $S_3(A_1, B) = 0.71$<br>$S_3(A_2, B) = 0.76$<br>$S_3(A_3, B) = 0.72$<br>$S_3(A_4, B) = 0.46$ | $A_2$ |
| Szmidt and Kacprzyk(2009) | The similarity measure based on the Hausdorff distance [8] | $S_5(A_1, B) = 0.72$<br>$S_5(A_2, B) = 0.76$<br>$S_5(A_3, B) = 0.72$<br>$S_5(A_4, B) = 0.46$ | $A_2$ |
| Jun Ye(2011) | Jun Ye's cosine similarity measure [16] | $S_7(A_1, B) = 0.9046$<br>$S_7(A_2, B) = 0.8832$<br>$S_7(A_3, B) = 0.8510$<br>$S_7(A_4, B) = 0.5033$ | $A_1$ |
| Hung and Wang(2012) | Hung and Wang's modified cosine similarity measure[17] | $S_8(A_1, B) = 0.8386$<br>$S_8(A_2, B) = 0.8765$<br>$S_8(A_3, B) = 0.8147$<br>$S_8(A_4, B) = 0.4616$ | $A_2$ |

| Author | The proposed method in this paper | $S(A_1,B) = 2.6628$<br>$S(A_2,B) = 2.6674$<br>$S(A_3,B) = 2.4442$<br>$S(A_4,B) = 1.4949$ | $A_2$ |

As what is shown in Table 4, the conclusions reached by the proposed method and by other methods are the same, except for the cosine similarity measure in Jun Ye(2011). This indicates that the method proposed in this paper is feasible in medical diagnosis.

## 6. Conclusion

IFS plays an important role in dealing with the uncertain and incomplete information which is characterized by the membership function and the non-membership function. For a MADM problem, the information of alternatives under different indicators is described in the form of IFN. This paper has defined a modified similarity measure of two IFSs based on cosine similarity and projection technology. This modified similarity measure considers not only the similarity degree of IFSs in the direction, but also the similarity degree in the length. In this way, more information can be included.

Furthermore, the proposed method in this paper has provided a calculation process which can be used to solve the MADM problems and medical diagnosis problems in IFS environment efficiently and effectively. And then the proposed method and some existing methods are compared and analyzed. The comparison result shows that the proposed method is effective and can identify the optimal scheme quickly.

The proposed method can be applied to not only IFS, but also IVIFS. But it has some limitation that it cannot be used for linguistic IFS. The membership degree and non-membership degree of IFS are expressed by numerical values, while the membership degree and non-membership degree of linguistic IFS are expressed by language. The proposed method can only handle numerical values and cannot handle linguistic variables. Therefore, it cannot be used to linguistic IFS.

In the future, the proposed method can be extended to the other uncertain and fuzzy environment. More other methods suitable to IFS and other fuzzy sets can be developed to solve the decision making problem. And other application area of the proposed method can be explored as well.

**Acknowledgements**

This research is supported by National Natural Science Fundation of China (Grant number: 72101165).